\documentclass[conference]{IEEEtran}
\usepackage[pdftex]{graphicx}

\let\labelindent\relax
\usepackage{enumitem}

\usepackage{amsfonts}

\graphicspath{{../pdf/}{../jpeg/}}
\DeclareGraphicsExtensions{.pdf,.jpeg,.png}
\usepackage{tabularx}
\usepackage{booktabs}
\usepackage[labelfont=bf,format=plain,justification=raggedright,singlelinecheck=false]{caption}
\usepackage[cmex10]{amsmath}
\usepackage{mathabx}
\usepackage{graphicx}
\usepackage{subcaption}
\usepackage{subfig}
\usepackage{floatrow}
\usepackage{array}
\usepackage{mdwmath}
\usepackage{caption}
\usepackage{mdwtab}
\usepackage{eqparbox}
\usepackage{url}
\usepackage{enumitem}
\usepackage{amsmath}
\usepackage{amssymb}
\usepackage{array}
\usepackage{multirow}
\usepackage{subcaption}
\usepackage{algorithm}
\usepackage[noend]{algpseudocode}
\usepackage{amsthm}

\begin{document}

\title{\LARGE
Pseudo-Feature Padding: A Lightweight Defense \\ Against False Data Injection in Power Grids
}

\author
{
     \IEEEauthorblockN{
         Farhin Farhad Riya\authorrefmark{1}, 
         Shahinul Hoque\authorrefmark{1}, 
         Yingyuan Yang\authorrefmark{2}, 
         Jinyuan Sun\authorrefmark{1},
         Kevin Tomsovic\authorrefmark{3}
     }
     
     \IEEEauthorblockA{
        \authorrefmark{1}Department of Electrical Engineering and Computer Science, University of Tennessee
     }
     \IEEEauthorblockA{
        \authorrefmark{2}Department of Computer Science, The University of Illinois at Springfield
     }
    \IEEEauthorblockA{
        \authorrefmark{3}Department of Electrical and Computer Engineering, Clemson University, South Carolina
     }
     \IEEEauthorblockA{ 
        friya@vols.utk.edu, 
        shoque@vols.utk.edu,
        yyang260@uis.edu, 
        jysun@utk.edu,
        ktomsov@clemson.edu    
    }
}

\maketitle

\begin{abstract}
Deep Neural Networks (DNNs) have achieved remarkable accuracy in various tasks, including their application in Cyber-Physical Systems (CPS) for detecting False Data Injection Attacks (FDIA) during critical operations. However, the unique infrastructure of CPS makes DNNs vulnerable to exploitation by attackers aiming to evade detection. Additionally, the distinct nature of CPS presents challenges for conventional defense mechanisms against FDIA. This paper proposes an innovative defense framework that strengthens DNNs against such attacks by introducing an additional input layer that performs padding in the input samples using pseudo-feature values derived from the input’s statistical distribution. This padding increases the input dimensionality in a randomized and data-aware manner, making adversarial attacks computationally infeasible due to the non-transferable nature of crafted perturbations and the unpredictability of the padded structure. Our method is lightweight, model-agnostic, and requires no modifications to the core architecture, making it highly deployable in real-world CPS settings. We evaluated our framework on critical power grid applications such as state estimation, using the IEEE 14-bus, 30-bus, 118-bus, and 300-bus systems. Experiments under adversarial settings demonstrate that our padding strategy significantly improves model robustness with negligible impact on performance, and effectively mitigates attacks that would otherwise bypass conventional defenses.

\end{abstract}

\begin{keywords}
Deep Neural Networks, False data injection attack, cyber-physical system.
\end{keywords}


\section{INTRODUCTION}
\vspace{-0.3em}
DNNs enhance accuracy and robustness in CPS such as power grids, transportation, and industrial control. However, the nature of their infrastructure exposes them to FDIA, which manipulates sensor and control signals with potentially severe consequences~\cite{liu2016masking}. Traditional defenses often fall short due to the intricate interdependencies in CPS. In power grids, for example, state estimation is vital for managing power flow and distribution but is vulnerable to FDIA that bypass anomaly detection and disrupt operations like real-time pricing~\cite{yu2015blind}. Despite extensive research and development of detection strategies, the unique constraints of power grid environments leave DNNs vulnerable to sophisticated perturbations \cite{falsedataattack}. 


In response to these vulnerabilities, this paper introduces a defense framework designed to mitigate the effect of FDIA against DNNs in CPS. Our approach involves the integration of an additional input layer that implements a padding strategy, augmented with pseudo-feature values derived directly from the statistical distribution of the input data \cite{goodfellow2014explaining}. Unlike prior padding-based defenses that rely on fixed values (e.g., zero-padding) \cite{li2021towards} \cite{riya2023mitigating}, our method dynamically generates pseudo-features by identifying low-importance input features through tree-based models and sampling new values from their fitted statistical distributions. This input-specific augmentation increases the dimensionality and complexity of the data in a way that makes adversarial perturbations significantly less transferable and more computationally expensive to generate. By introducing multiple padding combinations per training sample and randomizing the padding during inference, our method further enhances model robustness by increasing data diversity and adversarial uncertainty, without altering the core DNN architecture. This framework is designed to be lightweight and model-agnostic, requiring no fundamental changes to the core architecture of the existing DNN models, which facilitates easy deployment in real-world CPS settings. To demonstrate the efficacy of our proposed solution, we conduct extensive evaluations on the state estimation of the power grid and show that our padding strategy not only maintains the performance integrity of the DNNs but also significantly enhances their robustness. The paper also demonstrates that conventional defense techniques fall short in defending against these well-crafted FDIA samples. The key contributions of this paper can be summarized as follows:

\begin{itemize} 
  \item We propose a defense method that mitigates the effect of FDIA by padding the input samples with pseudo-features. This padding strategy not only increases the input dimensionality but also complicates the computation required to generate accurate FDIA samples.
  \item Our proposed framework can be easily adapted to the prevailing Machine Learning techniques that are utilized in FDIA detection in CPS.
  \item The framework requires no hardware changes or sensor redeployment, addressing the impracticality of securing all sensors in distributed CPS environments.
  \item Our proposed framework has a negligible accuracy drop compared to the baseline models.
  \item We validate the framework through simulation using the IEEE test system, including the IEEE 14-bus case, 30-bus case, 118-bus case, and 300-bus case. For every case, the favorable outcomes justify the proposed mechanism.
\end{itemize} 

\section{Related Work}
\vspace{-0.3em}
Since Liu et al. \cite{liu2009false} revealed SCADA system vulnerabilities to FDIA, extensive research has explored the threat due to the critical role of state estimation. Many works have studied stealthy FDIAs that exploit the structure of the Jacobian matrix to evade residual-based detectors \cite{kosut2011malicious, kim2013topology}, with some attacks succeeding even with partial system knowledge \cite{rahman2012false, srivastava2013modeling}. Other studies \cite{yan2016power, jiongcong2016impact} emphasize the risk of FDIA to system stability and operations. The impact of random and structured bad data on estimation was further analyzed in \cite{tajer2014energy}, highlighting persistent security concerns.
To defend against FDIA, early strategies focused on enhancing measurement security. For example, Huang et al.~\cite{bi2011defending} introduced an adaptive cumulative sum method for rapid detection, while others incorporated PMU measurements synchronized with GPS signals. More recently, machine learning (ML) has emerged as a computationally efficient, hardware-free alternative~\cite{ozay2016machine, esmalifalak2014detecting}. ML-based approaches include supervised and semi-supervised detection using Gaussian models~\cite{ozay2016machine}, and anomaly detection via unsupervised and SVM-based classifiers~\cite{esmalifalak2014detecting}. Deep learning models such as RNNs with LSTM cells and convolutional networks have demonstrated strong performance against FDIA~\cite{mohammad2018detecting, JQ2018Online}. However, these models remain vulnerable to adversarial perturbations. Szegedy et al.~\cite{szegedy2013intriguing} first revealed that DNNs can be manipulated using carefully crafted inputs. Since then, several attack techniques have been proposed, including FGSM~\cite{goodfellow2014explaining}, FGM~\cite{rozsa2016adversarial}, iterative methods~\cite{kurakin2016adversarial}, and DeepFool~\cite{moosavi2016deepfool}.

Additionally, defense mechanisms such as adversarial training~\cite{kurakin2016adversarial}, model distillation~\cite{papernot2016distillation}, adversarial input detection~\cite{xu2017feature}, and input reconstruction~\cite{gu2014towards} have been explored. More recent approaches include deep reinforcement learning for adaptive FDIA detection~\cite{deepReinforcementLearningFDIA}, multi-mode attack detection strategies~\cite{multiAttackModeDetection}, and GAN-GRU-based data balancing to mitigate class imbalance~\cite{GANBasedDataBalancing}. However, most of these methods require substantial modifications to the training pipeline or model architecture. In contrast, our proposed framework operates as a lightweight preprocessing step that preserves the original model. Moreover, we show that existing techniques such as adversarial training and distillation perform poorly in CPS-like settings due to the constraints. Prior padding-based defenses~\cite{li2021towards, riya2023mitigating} improve robustness via zero-padding but are limited by their fixed padding values, which can be exploited in white-box scenarios. Our method overcomes this limitation by introducing randomized, data-aware padding that increases input complexity and reduces attack transferability.

\begin{figure*}[htbp]
    \centering
    \begin{subfigure}{0.32\columnwidth}
        \centering
        \includegraphics[width=\columnwidth,height=3.3cm]{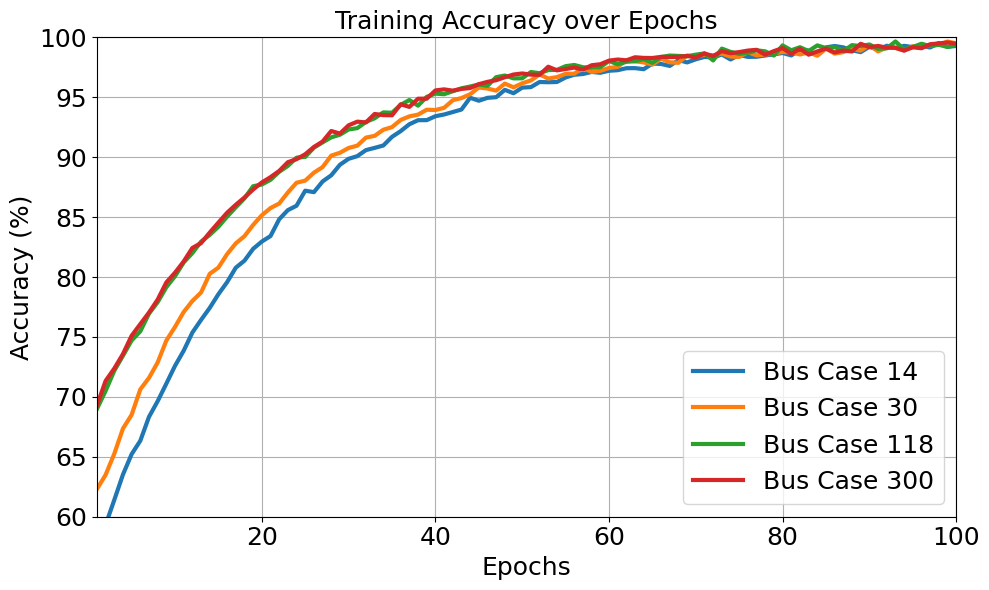}
        \caption{Model training accuracy}
        \label{fig:fig1sub1}
    \end{subfigure}
    \hfill
    \begin{subfigure}{0.33\columnwidth}
        \centering
        \includegraphics[width=0.8\columnwidth, height=3.3cm]{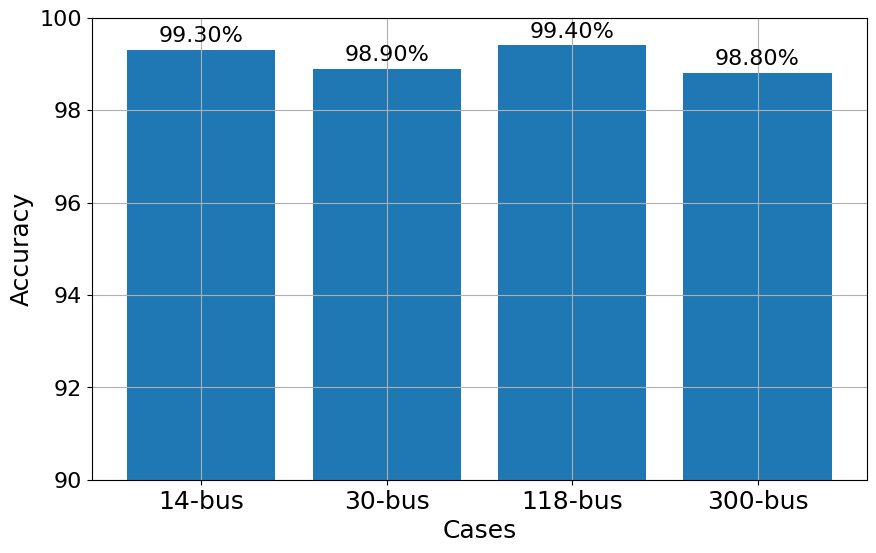}
        \caption{Model detection accuracy}
        \label{fig:fig1sub2}
    \end{subfigure}
    \hfill
    \begin{subfigure}{0.32\columnwidth}
        \centering
        \includegraphics[width=0.8\columnwidth, height=3.3cm]{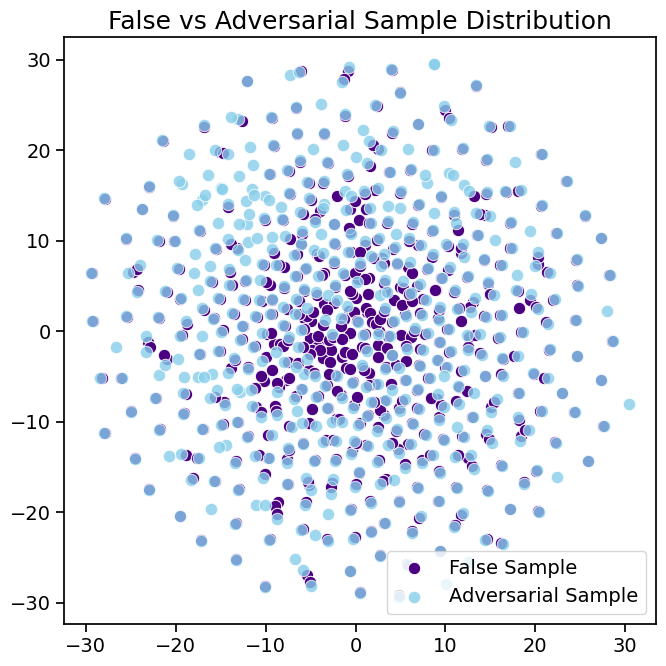}
        \caption{Manifold of the false and adversarial samples}
        \label{fig:figmanifold}
    \end{subfigure}
    \caption{Experimental setup: Baseline model performance under no attack and the manifold of false and adv-measurements}
    \label{fig:fig1}
\vspace{-0.5em}
\end{figure*}

\section{Background}
\vspace{-0.3em}
\textbf{Notations:} $m$ and $n$ denote the number of measurements and state variables. The Jacobian matrix $H \in \mathbb{R}^{m \times n}$ maps the state vector $x \in \mathbb{R}^n$ to the measurement vector $z \in \mathbb{R}^m$. Measurement noise is represented by $e$, $\hat{x}$ is the estimated state, and $W$ is the noise covariance matrix. \textit{Bias L$_2$-Norm} captures the estimation shift from adversarial perturbations, while \textit{Valid L$_2$-Norm} quantifies perturbation strength.


\vspace{-0.7em}
\subsection{State Estimation in Power System}
\vspace{-0.3em}
State estimation in DC power systems is fundamental for understanding the dynamics and integrity of the network. It utilizes the linear model, \(\displaystyle z = Hx + e\),
where $x = (x_1, x_2, \ldots, x_n)^T$ denotes the state variables of the power system, $z = (z_1, z_2, \ldots, z_m)^T$ are the sensor measurements, and $H$ is the $m \times n$ Jacobian matrix relating measurements to the states. The error vector $e = (e_1, e_2, \ldots, e_m)^T$ accounts for measurement inaccuracies. The system is typically over-determined ($m > n$), and the state estimation is often performed via a weighted least squares approach:
\begin{equation}
   \hat{x} = (H^TWH)^{-1}H^TWz \label{eq:2}
\end{equation}
Here, $W$ is a diagonal weight matrix derived from the inverse variances of the measurement errors, optimizing the estimation accuracy by minimizing the variance of the estimation error. This formulation ensures robust state estimation even in the presence of measurement redundancy and potential data inaccuracies, ensuring secure and efficient power grid operations.
\textit{Bad Data Detection:}
Sensor measurements in state estimation can be compromised by faults, misconfigurations, or malicious actions. To detect such errors, the L$_{2}$-norm of the residual $|z - H\hat{x}|$ is evaluated, where $\hat{x}$ is the estimated state. If this norm exceeds a threshold $\tau$, the data is flagged as bad. Assuming Gaussian noise, the squared residual follows a chi-squared distribution with $\nu = m - n$ degrees of freedom, and $\tau$ is set via hypothesis testing at significance level $\alpha$.


\subsection{False Data Injection Attack}
FDIA enables attackers to manipulate state estimations in power systems by introducing perturbed measurements. If an attacker knows the system's matrix $H$, they can generate a perturbation vector $a = [a_1, a_2, \ldots, am]^T$, leading to polluted measurements $z_a = z + a$. This attack aims to make polluted measurements indistinguishable from legitimate ones by ensuring that the error does not exceed a detection threshold $\tau$. The relationship between the compromised and estimated states can be mathematically represented as:
\begin{equation}
   ||z_a - H\hat{x}_{bad}|| = ||z + a - H(\hat{x} + c)|| \label{eq:4a}
\end{equation}
where $\hat{x}_{bad}$ and $\hat{x}$ denote the estimated states from $z_a$ and $z$ respectively. Simplifying, we find:
\begin{equation}
   ||z - H\hat{x} + (a - Hc)|| \leq \tau \label{eq:4b}
\end{equation}

To construct an effective attack vector $a$, the attacker uses the projection matrix $P = H(H^T H)^{-1}H^T$ and defines $B = P - I$. For the attack to remain undetected, $a$ must satisfy $Ba = 0$. In practical scenarios where only a subset of measurements can be compromised and the attackers aim to construct the vector $a$ such that its non-zero components correspond to the compromised measurements, while still satisfying $Ba = 0$.

\begin{figure} [htbp]
    \centering
    \includegraphics[width=0.7\columnwidth, height=3cm]{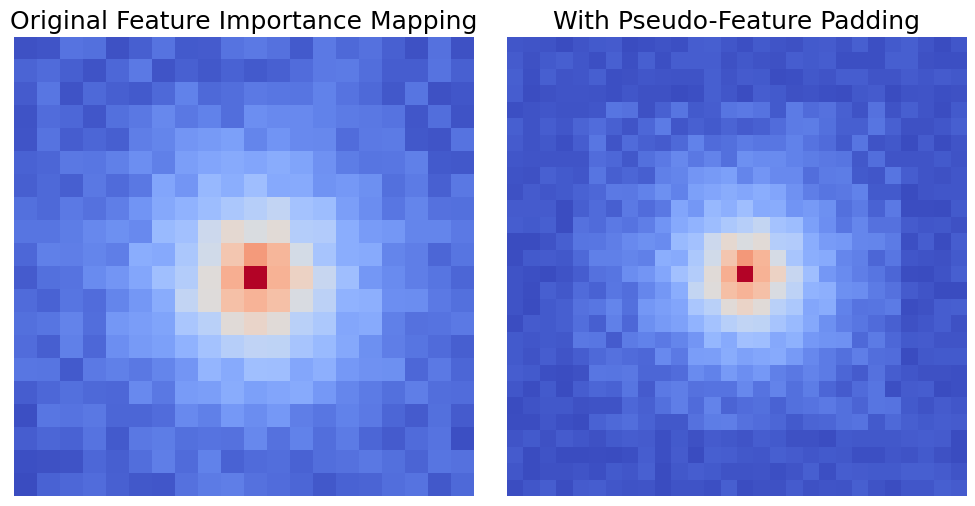}
    \caption{Pseudo-Feature Padding input modification}
    \label{fig:framework}
\vspace{-0.5em}
\end{figure}

Our goal is to defend a DNN-based anomaly detector \( f_\theta(\cdot) \) against such structured perturbations. We define a transformation \( \Phi(z) = \text{concat}(z, P_{\text{pad}}) \), where \( P_{\text{pad}} \in \mathbb{R}^p \) is a pseudo-feature vector generated by sampling from the distributions of low-importance features identified via a tree-based model. This padding operation increases the input dimensionality and introduces structural randomness, which modifies the projection matrix used in detection and breaks the attacker's stealth condition \( Ba = 0 \).

\section{\textbf{Proposed Pseudo-feature Padding}}
In response to the vulnerabilities exposed by FDIA and the criticality of accurate state estimation in power systems, our framework introduces a pseudo-feature padding technique.

\subsubsection{\textbf{Overview of the Framework}}

The proposed framework enhances robustness against adversarial manipulations by appending statistically derived pseudo-features to each input, thereby increasing its dimensionality and introducing structured randomness. Formally, for any input measurement \( \tilde{z} \in \{ z, z' \} \), where \( z \) is a clean sample and \( z' = z + a \) is an adversarially perturbed version, we define the padded input as:
\begin{equation}
    z_2 = \Phi(\tilde{z}) = \text{concat}(\tilde{z}, P_{\text{pad}})
    \label{eq:padding_concat}
\end{equation}

Here, \( P_{\text{pad}} \) denotes a pseudo-feature vector generated from the statistical distribution of low-importance features identified using a decision tree-based analysis. While traditional FDIA assumes additive perturbations designed to satisfy the stealth condition \( Ba = 0 \), our input transformation increases the input dimension and modifies the detection space. As a result, previously stealthy attack vectors are no longer aligned with the null space of the transformed system, reducing attack effectiveness and transferability.

During training, pseudo-features are concatenated to clean samples \( z \), generating multiple augmented variants per sample. Each training sample undergoes feature importance ranking to identify low-saliency features \( f_i \), from which synthetic values \( f' \) are sampled to form \( P_{\text{pad}} \). A padding size \( p \) controls how many variants are generated per sample, leading to a dataset expansion from \( Z \) to \( (p+1)Z \) samples.

At inference, the model may encounter either a clean sample \( z \) or an adversarially perturbed sample \( z' \). In both cases, a fresh pseudo-feature vector \( P_{\text{pad}} \) is sampled and concatenated to the observed input, forming \( z_2 \). This consistent application of padding ensures robustness while preserving performance on benign inputs.

\begin{algorithm}[htbp]
\caption{Pseudo-feature Padding with Per-Sample Importance}
\label{alg:alo1}
\scriptsize
\begin{algorithmic}[1]
\State \textbf{Input:} Dataset $D = \{(x^{(i)}, y^{(i)})\}$, tree-based model, distribution fitting method, threshold $\theta$, weights $W$, bias $b$
\State \textbf{Output:} Trained model with pseudo-padded input robustness

\For{each feature $f_j$ in $x$}
    \State Fit distribution $Dist_j$ over all values of $f_j$ in $D$ \Comment{Used for sampling}
\EndFor

\For{each sample $(x^{(i)}, y^{(i)}) \in D$}
    \State Compute feature importance scores $S_j^{(i)}$ for $x^{(i)}$ using the tree-based model
    \State $F_{\text{less}}^{(i)} \gets \{f_j : S_j^{(i)} < \theta\}$ \Comment{Less important features for this sample}
    \State Sample pseudo-features $\{p_j^{(i)}\}$ from $Dist_j$ for $f_j \in F_{\text{less}}^{(i)}$
    \State $x^{(i)}_{\text{new}} \gets$ concatenate($x^{(i)}$, $\{p_j^{(i)}\}$)
\EndFor

\State Train DNN using $(x^{(i)}_{\text{new}}, y^{(i)})$ to minimize loss with respect to $W$, $b$

\For{each inference sample $(x^{(\text{test})}, y^{(\text{test})})$}
    \State Compute $S_j^{(\text{test})}$ for $x^{(\text{test})}$
    \State $F_{\text{less}}^{(\text{test})} \gets \{f_j : S_j^{(\text{test})} < \theta\}$
    \State Sample pseudo-features $\{p_j^{(\text{test})}\}$ from $Dist_j$ for $f_j \in F_{\text{less}}^{(\text{test})}$
    \State $x^{(\text{test})}_{\text{new}} \gets$ concatenate($x^{(\text{test})}$, $\{p_j^{(\text{test})}\}$)
    \State Predict $\hat{y}^{(\text{test})} \gets$ DNN($x^{(\text{test})}_{\text{new}}$)
    \State Check for FDIA anomalies
\EndFor
\end{algorithmic}
\end{algorithm}

Algorithm~\ref{alg:alo1} outlines the proposed framework. For each input sample \( x^{(i)} \) in dataset \( D \), a tree-based model ranks feature importance scores \( S_j^{(i)} \), and features below a threshold \( \theta \) are identified as less important. Statistical distributions are pre-fitted to each feature across the entire dataset, and pseudo-features are sampled from these distributions for the low-importance features of each sample. The sampled pseudo-features are concatenated with \( x^{(i)} \) to form the augmented input \( x^{(i)}_{\text{new}} \), increasing input dimensionality and structural diversity. The DNN is trained on these padded inputs, optimizing weights \( W \) and bias \( b \) while enhancing robustness to adversarial perturbations.

\subsubsection{\textbf{Robustness of the Pseudo-Feature Padding}}
The attack vector \( a \) is designed such that \( B a = 0 \), where \( B = P - I \) and \( P = H(H^T H)^{-1}H^T \) is the projection matrix onto the column space of \( H \). With the introduction of pseudo-feature padding \( P_{\text{pad}} \), the effective system matrix becomes \( H' = [H \mid P_{\text{pad}}] \), modifying the projection matrix to \( P' = H'(H'^T H')^{-1}H'^T \). This transforms the residual matrix to \( B' = P' - I \), and the adversarial condition becomes \( B' a' = 0 \), where \( a' \) includes perturbations in both original and padded features.

\noindent\textbf{Lemma 1.}
Let \( B = P - I \) and \( B' = P' - I \) be the projection residual matrices before and after padding. Then \( \mathcal{N}(B) \neq \mathcal{N}(B') \); that is, the null space structure changes under pseudo-feature padding.

\noindent\textbf{Proof Sketch.}
Let \( H \in \mathbb{R}^{m \times n} \) be the system Jacobian and \( P = H(H^T H)^{-1} H^T \). Pseudo-feature padding constructs \( H' = [H \,|\, P_{\text{pad}}] \), where \( P_{\text{pad}} \in \mathbb{R}^{m \times p} \) contains vectors sampled from the fitted distributions of low-importance features identified via a tree-based model. This increases the column space dimensionality, shifting the null space of the projection matrix \( P' = H'(H'^T H')^{-1} H'^T \). Consequently, an attack vector \( a \) satisfying \( Ba = 0 \) will generally fail to satisfy \( B'a' = 0 \), as \( \mathcal{N}(B') \) is no longer aligned with \( \mathcal{N}(B) \). Additionally, random sampling of \( P_{\text{pad}} \) at inference makes reconstructing \( H' \) infeasible, thus preventing transferable attacks. This change in null space complicates the attacker’s optimization problem. Since FDIA relies on precise structure in \( H \), padding introduces structural uncertainty that increases adversarial difficulty while preserving model accuracy.

\subsubsection{\textbf{Impact of Less Important Features on Model Accuracy}}
Less important features are characterized by low importance scores, which minimally contribute to the accuracy of machine learning models. Breiman points out that these features have negligible effects on model performance when altered or removed \cite{breiman2001random}. Empirical studies confirm that excluding these features does not significantly degrade, and may even improve, model accuracy by reducing complexity and overfitting \cite{gregorutti2017correlation, guyon2003introduction, kuhn2013applied}. In light of these studies, the impact of our pseudo-feature padding on model accuracy is expected to be very negligible. Additional evaluation results provided in subsequent sections further justify the statement. 

Figure \ref{fig:framework} illustrates the original input's feature importance heatmap (left) and its augmentation with pseudo-features resembling low-importance areas (right). This visualization demonstrates that the distribution of important features within the input sample remains intact, ensuring that the augmentation does not distort the model's decision-making process.

\begin{table}[htbp]
\centering
\caption{Performance Comparison (Attack sample defection accuracy, Bias L$_2$-Norm and Valid L$_2$-Norm) of Different Defense Techniques}
\label{tab:comp}
\begin{tabular}{|c|c|c|c|c|}
\hline
\textbf{\small Defense} & \textbf{\small Case} & \textbf{\small Accuracy} & \textbf{\small Bias L$_{2}$} & \textbf{\small Valid L$_{2}$} \\ \hline
\multirow{4}{*}{\small Plain DNN} & \small 14-bus & \small 35.1\% & \small 70.7 & \small 119.8 \\ \cline{2-5} 
                          & \small 30-bus & \small 30.4\% & \small 87.2 & \small 112.2 \\ \cline{2-5}
                          & \small 118-bus & \small 39.7\% & \small 64.5 & \small 114.0 \\ \cline{2-5}
                          & \small 300-bus & \small 40.0\% & \small 109 & \small 179.0 \\ \hline
                          
\multirow{4}{*}{\small Adversarial Distilled} & \small 14-bus & \small 45.7\% & \small 65.2 & \small 105.5 \\ \cline{2-5} 
                          & \small 30-bus & \small 42.6\% & \small 63.4 & \small 100.8 \\ \cline{2-5}
                          & \small 118-bus & \small 40.2\% & \small 59.8 & \small 110.7 \\ \cline{2-5}
                          & \small 300-bus & \small 48.3\% & \small 55.1 & \small 120.2 \\ \hline
\multirow{4}{*}{\small Adversarial Training} & \small 14-bus & \small 62.0\% & \small 50.6 & \small 113.7 \\ \cline{2-5} 
                           & \small 30-bus & \small 58.7\% & \small 49.0 & \small 109.9 \\ \cline{2-5}
                           & \small 118-bus & \small 60.1\% & \small 47.5 & \small 120.2 \\ \cline{2-5}
                           & \small 300-bus & \small 64.4\% & \small 44.2 & \small 125.6 \\ \hline
\multirow{4}{*}{\small Zero Padding ~\cite{ riya2023mitigating} } 
    & \small 14-bus & \small 93.5\% & \small 159.9 & \small 44.2 \\ \cline{2-5} 
    & \small 30-bus & \small 93.1\% & \small 168.5 & \small 51.0 \\ \cline{2-5}
    & \small 118-bus & \small 93.7\% & \small 167.2 & \small 58.7 \\ \cline{2-5}
    & \small 300-bus & \small 95.0\% & \small 163.8 & \small 45.8 \\ \hline

\multirow{4}{*}{\small Pseudo-feature} 
    & \small 14-bus & \small 94.2\% & \small 168.7 & \small 55.1 \\ \cline{2-5} 
    & \small 30-bus & \small 92.0\% & \small 172.3 & \small 48.4 \\ \cline{2-5}
    & \small 118-bus & \small 93.4\% & \small 174.9 & \small 49.0 \\ \cline{2-5}
    & \small 300-bus & \small 96.1\% & \small 174.7 & \small 48.5 \\ \hline
\end{tabular}
\end{table}

\begin{figure}[htp]
    \centering
    \begin{subfigure}{0.48\columnwidth}
        \centering
        \includegraphics[width=\textwidth, height=2.5cm]{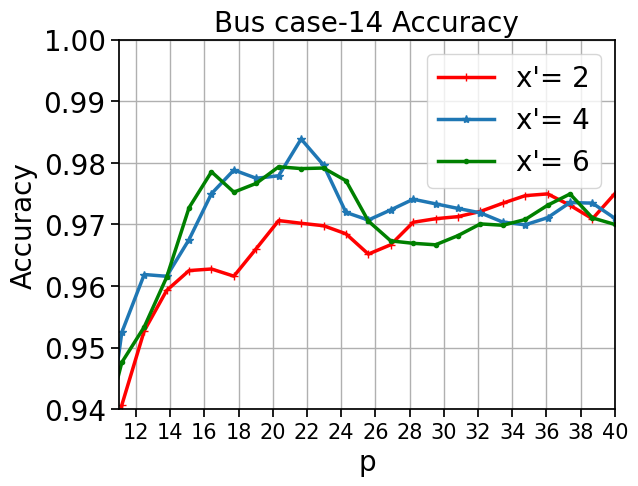}
        \vspace{-1em}   
        \label{fig:fig2sub1}
    \end{subfigure}
    \begin{subfigure}{0.48\columnwidth}
        \centering
        \includegraphics[width=\textwidth, height=2.5cm]{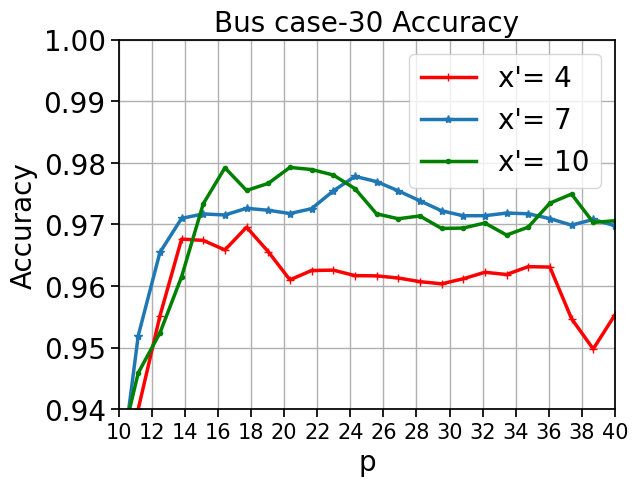}
        \vspace{-1em}   
        \label{fig:fig2sub2}
    \end{subfigure}

    \begin{subfigure}{0.48\columnwidth}
        \centering
        \includegraphics[width=\textwidth, height=2.5cm]{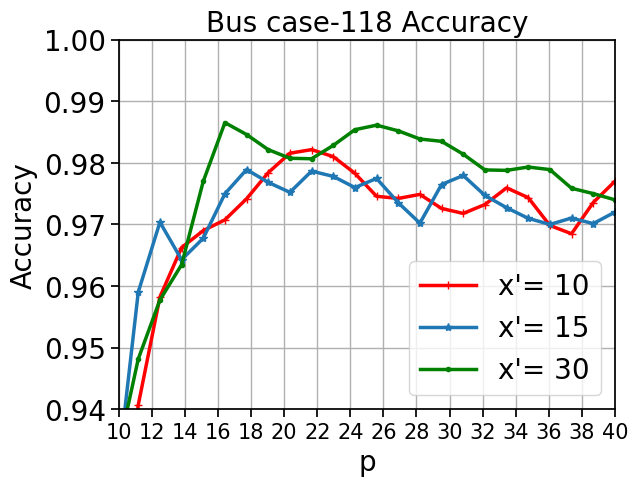}
        \vspace{-1em}   
        \label{fig:fig2sub3}
    \end{subfigure}
    \begin{subfigure}{0.48\columnwidth}
        \centering
        \includegraphics[width=\textwidth, height=2.5cm]{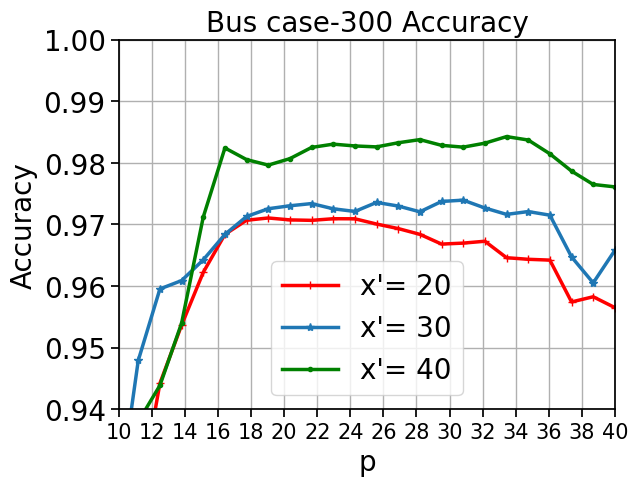}
        \vspace{-1em}   
        \label{fig:fig2sub4}
    \end{subfigure}
\caption{Attack detection accuracy for bus case-14, case-30, case-118 and case-300 with increasing padding size $p$}
\vspace{-2pt}
\label{fig:fig2}    
\end{figure}

\section{Evaluation}

\subsection{Experimental Setup}

\subsubsection{\textbf{Dataset}}
The behavior of the power systems was simulated using a DC power flow model in MATPOWER (MATLAB) to construct the \( H \) matrix linking state variables to meter measurements. The dataset contains 40,000 samples across multiple IEEE test cases, evenly split between normal and false data for binary classification (\( 0 \) = normal, \( 1 \) = false). Robustness was evaluated by varying the number of compromised meters; notably, compromising just 4\% significantly reduced detection accuracy, highlighting system vulnerability. This setup offers a realistic benchmark for assessing our framework under adversarial conditions.

\begin{figure}[htp]
    \centering
    \begin{subfigure}{0.48\columnwidth}
        \centering
        \includegraphics[width=\textwidth, height=2.5cm]{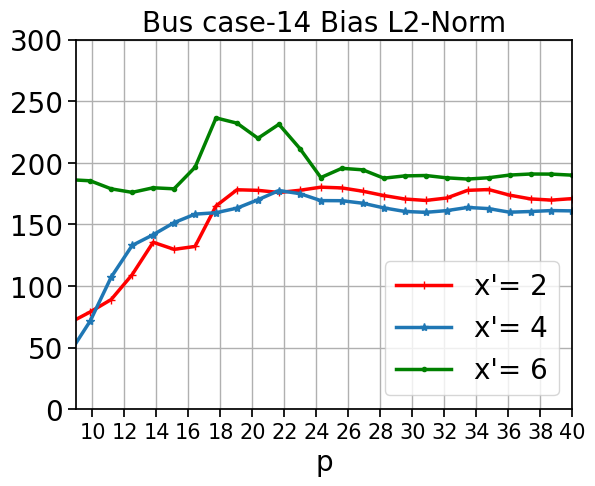}
        \vspace{-1em}   
        \label{fig:fig3sub1}
    \end{subfigure}
    \begin{subfigure}{0.48\columnwidth}
        \centering
        \includegraphics[width=\textwidth, height=2.5cm]{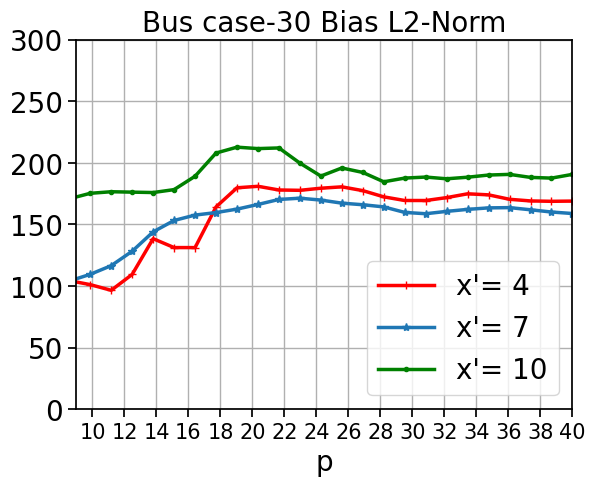}
        \vspace{-1em}   
        \label{fig:fig3sub2}
    \end{subfigure}
    \begin{subfigure}{0.48\columnwidth}
        \centering
        \includegraphics[width=\textwidth, height=2.5cm]{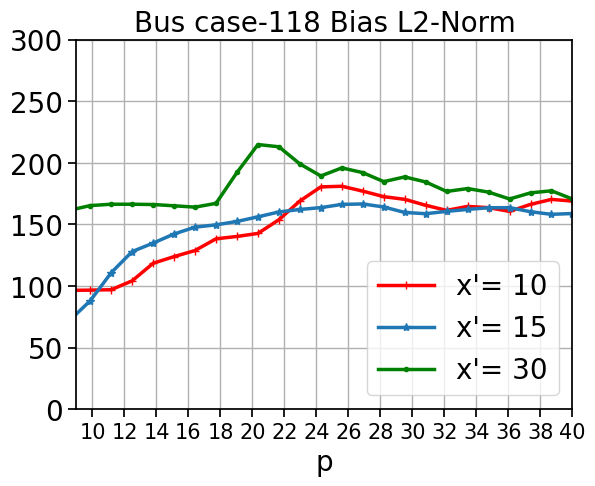}
        \vspace{-1em}   
        \label{fig:fig3sub3}
    \end{subfigure}
    \begin{subfigure}{0.48\columnwidth}
        \centering
        \includegraphics[width=\textwidth, height=2.5cm]{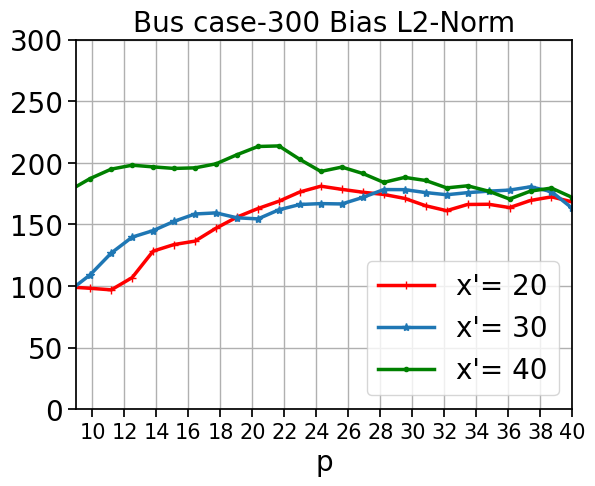}
        \vspace{-1em}   
        \label{fig:fig3sub4}
    \end{subfigure}
\caption{Bias L$_2$-Norm for bus case-14, case-30, case-118 and case-300 with increasing padding size $p$}
\vspace{-2pt}
\label{fig:fig3}    
\end{figure}

\subsubsection{\textbf{Defense Model}}
In the simulation, we use a feed-forward DNN, denoted as network \( F \), comprising four fully connected layers with ReLU activations, dropout regularization, and a sigmoid output layer for binary classification. The input layer size varies with the selected padding size, which determines the number of pseudo-features. Training is performed using the Adam optimizer (learning rate 0.001), batch size 128, for 100 epochs.

\subsubsection{\textbf{Attack Sample Generation}}
Detection accuracy was evaluated using 1000 adversarial samples generated via the iterative projection framework in \cite{kurakin2016adversarial, madry2017towards}, under power system constraints. The number of compromised meters was varied per test case. A \textit{Collective-Pattern Attack} was also considered, where informed attackers target randomized models to evade detection. As shown in Figure~\ref{fig:figmanifold}, adversarial data closely aligns with false data, enabling it to bypass plain DNNs due to system-constrained manifolds that we discussed in the Background sections. Adaptive and partial-information attacks are left for future work, though robustness is expected to hold under the non-deterministic padding scheme.

\subsubsection{\textbf{Attacker’s Knowledge}}
We consider a white-box threat model where the attacker knows the model and training data but not the pseudo-feature padding. Since pseudo-features are randomly sampled at inference, the input structure varies unpredictably, reducing the success and transferability of adversarial perturbations.

\subsubsection{\textbf{Evaluation Metrics}}
Model performance was evaluated using three metrics: Attack Detection Accuracy, which measures the system's ability to identify malicious inputs; Bias L$_2$-Norm, assessing the error from adversarial perturbations; and Valid L$_2$-Norm, quantifying the size of adversarial perturbations. Success from an attacker's view is indicated by lower accuracy, smaller Bias L$_2$-Norm, and larger Valid L$_2$-Norm.

\begin{figure}[t]
    \centering
    \begin{subfigure}{0.48\columnwidth}
        \centering
        \includegraphics[width=\textwidth, height=2.5cm]{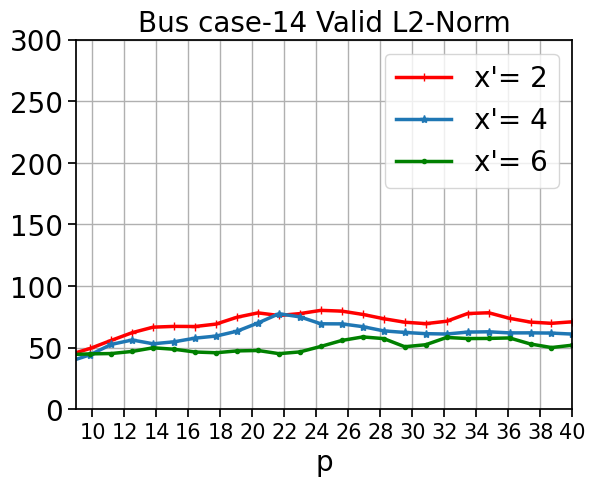}
        \vspace{-1em}   
        \label{fig:sfig4ub1}
    \end{subfigure}
    \begin{subfigure}{0.48\columnwidth}
        \centering
        \includegraphics[width=\textwidth, height=2.5cm]{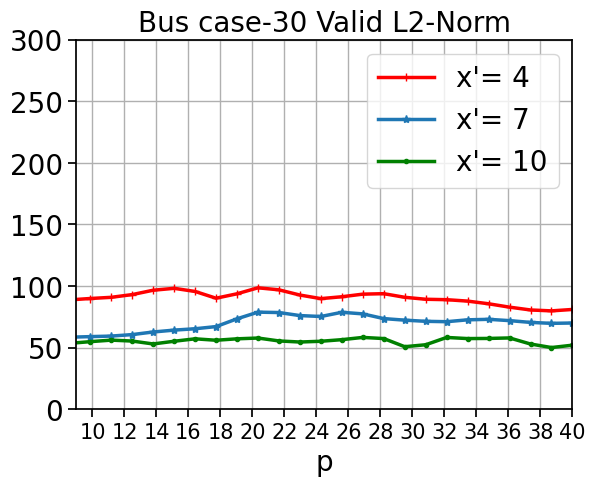}
        \vspace{-1em}   
        \label{fig:fig4sub2}
    \end{subfigure}
    \begin{subfigure}{0.48\columnwidth}
        \centering
        \includegraphics[width=\textwidth, height=2.5cm]{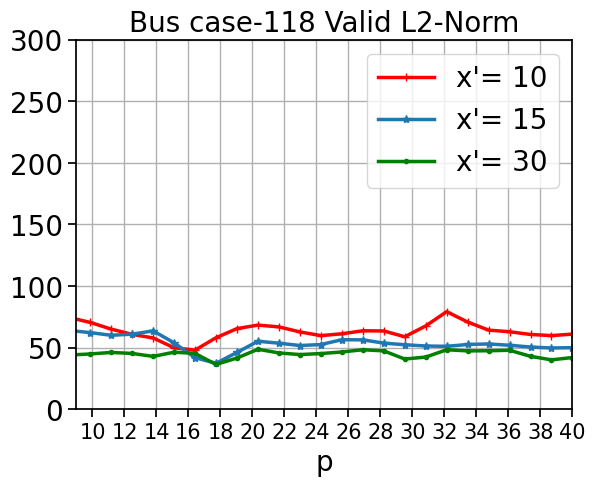}
        \vspace{-1em}   
        \label{fig:fig4sub3}
    \end{subfigure}
    \begin{subfigure}{0.48\columnwidth}
        \centering
        \includegraphics[width=\textwidth, height=2.5cm]{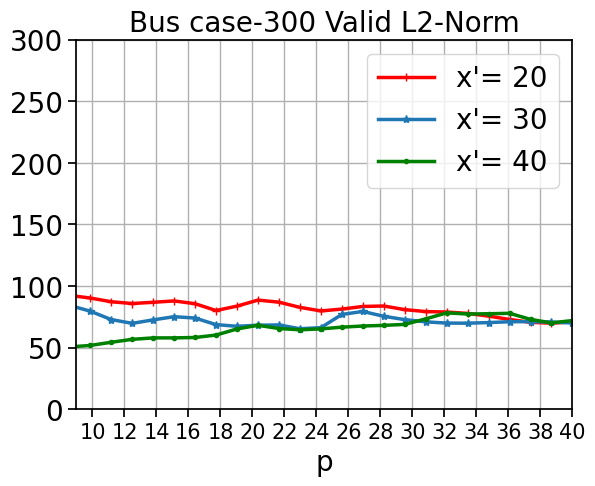}
        \vspace{-1em}   
        \label{fig:fig4sub4}
    \end{subfigure}
\caption{Valid L$_2$-Norm for bus case-14, case-30, case-118 and case-300 with increasing padding size $p$}
\vspace{-1em}
\label{fig:fig4}    
\end{figure}

\begin{table}[htbp]
\centering
\footnotesize
\caption{Computational Overhead Comparison}
\begin{tabular}{|l|c|c|c|c|c|}
\hline
\textbf{Method} & \textbf{Train (s)} & \textbf{Infer (ms)} & \textbf{\#Params} & \textbf{Steps} & \textbf{Overhead} \\
\hline
Plain DNN & 10 & 0.8 & 145k & None & Low \\
Adv Train & 35 & 1.1 & 145k & Grad gen & High \\
Adv Distillation & 20 & 1.0 & 365k & T-S distill & Moderate \\
Pseudo-Feature & 35 & 0.9 & 437k & Gen+Pad & Low-Mod \\
\hline
\end{tabular}
\vspace{-0.5em}
\label{tab:overhead}
\end{table}

\subsection{Results}
\subsubsection{\textbf{Baseline Results}}
Figure \ref{fig:fig1sub1} shows the accuracy achieved by the plain DNN model on four bus cases during training. All cases demonstrate high accuracy, indicating strong training performance across different system complexities. Figure \ref{fig:fig1sub2} illustrates the detection accuracy of the plain DNN model under no attack. The detection accuracy was validated with similar attack samples on which the model was trained. This will give us a baseline accuracy for further evaluation.

\subsubsection{\textbf{Comparison with the State-of-art Defense techniques}}
Table~\ref{tab:comp} compares the performance of various defense techniques against adversarial attacks, using Attack Detection Accuracy, Bias L$_2$-Norm, and Valid L$_2$-Norm as metrics. Results indicate that the Adversarial Training technique generally outperforms Plain DNN and Adversarial Distilled methods (avg. distillation temp 20-100) across various bus cases. Specifically, Adversarial Training achieves higher detection accuracies and consistently lower Bias L$_2$ values, indicating a stronger resistance to attacks. However, the Valid L$_2$ values are generally higher for Adversarial Training, suggesting that while it detects attacks more effectively, the intensity of the adversarial perturbations it faces might be greater, possibly due to attackers targeting these more robust defenses more aggressively.
Despite moderate improvements, these defenses remain insufficient in CPS settings. Distillation assumes small perturbations, which is ineffective for FDIA, where attackers prioritize stealth over minimality. Adversarial training is computationally intensive and performs poorly against iterative attacks. Detection methods fail because adversarial inputs in FDIA share the same manifold as normal false data, making distribution-based separation unreliable under power system constraints. While zero-padding performs well as our proposed method, its fixed padding strategy can be exploited in white-box scenarios where the attacker is aware of the zero values, undermining its robustness.

\subsubsection{\textbf{Results of the proposed framework}}
Figure~\ref{fig:fig2} illustrates the accuracy trends of the proposed pseudo-feature padding technique across four standard IEEE bus systems. For each case, the padding size $p$ is varied, and accuracy is evaluated for different numbers of compromised meters $x'$. Across all cases, accuracy improves with increasing padding size up to a saturation point, after which marginal drops or fluctuations are observed. Notably, larger systems (e.g., 118-bus and 300-bus) achieve better stability and higher peak accuracy, supporting the scalability of the framework. Figure~\ref{fig:fig3} illustrates the Bias L$_2$-Norm values as a function of padding size $p$ for all the bus cases under varying compromised meters. As observed, increasing the padding size initially results in a sharp increase in the Bias L$_2$-Norm, which then stabilizes or fluctuates slightly for larger $p$. The larger Bias L$_2$-Norm value indicates the less successful attack examples, meaning better model detection. Similarly, figure~\ref{fig:fig4} illustrates a general trend where smaller values of $p$ tend to result in higher L$_2$ norms, indicating a stronger adversarial impact. As $p$ increases, the injected perturbations are more efficiently neutralized, reflected by reduced Valid L$_2$-Norm values. Notably, the results suggest the padding strategy limits the transferability of adversarial examples, thus decreasing the effectiveness of the attacks. These findings highlight the role of larger pseudo-feature sets in enhancing defense robustness by reducing the effective perturbation strength as perceived in the state estimation layer.

\subsubsection{\textbf{Computational overhead}}
Table~\ref{tab:overhead} compares the computational overhead of different defense methods. While our pseudo-feature framework introduces a pseudo-feature generating step and a padding step that increases training time compared to the plain model, it maintains low inference cost and moderates overall overhead.

\section{Conclusion}
\vspace{-0.3em}
FDIA continues to pose a serious threat to deep learning-based detection frameworks in CPS, like power grids, especially in environments where securing physical-level sensors is infeasible and often dependent on specific hardware configurations. In this paper, we propose a lightweight, model-agnostic defense mechanism that reconstructs (pads) input samples, adding pseudo-features derived from the statistical distribution of less important input features. This strategy introduces padding the inputs with pseudo-features and increases complexity, which significantly reduces the effectiveness of adversarial perturbations without compromising model performance. We validate our framework against strong adversarial attacks across multiple IEEE test systems and demonstrate its robustness, minimal computational overhead, and ease of integration into existing DNN-based FDIA detection pipelines. Our results affirm the potential of pseudo-feature padding as a practical and effective defense solution for real-world CPS applications.

\vspace{-0.5em}
{ 
\bibliographystyle{IEEEtran}
\bibliography{IEEEabrv,reference}
}

\noindent\textbf{Acknowledgment.}
This work was supported by the US National Science Foundation (NSF) under grant CNS-2038922.

\end{document}